\documentclass{article}

\PassOptionsToPackage{numbers, compress}{natbib}

\usepackage[utf8]{inputenc} 
\usepackage[T1]{fontenc}    
\usepackage{hyperref}       
\usepackage{url}            
\usepackage{booktabs}       
\usepackage{amsfonts}       
\usepackage{nicefrac}       
\usepackage{microtype,multirow}      
\usepackage{bm,amsmath,amssymb}
\usepackage{xspace}
\usepackage{graphicx}
\usepackage{color}
\usepackage{natbib}

\usepackage{colortbl}
\definecolor{light_gray}{RGB}{170,170,170}

\usepackage[utf8]{inputenc} 
\usepackage[T1]{fontenc}    
\usepackage{hyperref}       
\usepackage{url}            
\usepackage{booktabs}       
\usepackage{amsfonts}       
\usepackage{nicefrac}       
\usepackage{microtype}      
\usepackage{longtable,lscape}

\usepackage{subcaption}

\usepackage{wrapfig}

\newcommand{\hide}[1]{}

\usepackage{enumitem}

\usepackage{bm}

\newcommand{\bb}{\mathbf{b}}

\newcommand{\rbb}{\mathbf{r}}
\newcommand{\sbb}{\mathbf{s}}

\newcommand{\calL}{\mathcal{L}}

\newcommand{\calY}{\mathcal{Y}}

\usepackage{color}

\usepackage{tikz}
\usetikzlibrary{intersections}
\usepackage{booktabs}
\usetikzlibrary{calc}
\usetikzlibrary{decorations.pathreplacing}
\usetikzlibrary{shapes,arrows,positioning}
\usetikzlibrary{chains, positioning}

\usepackage{enumitem}
\usepackage{amsmath}
\usepackage{amssymb}

\usepackage{graphicx}
\usepackage{rotating}
\usepackage{array}
\usepackage{algorithm}
\usepackage{algorithmic}
\usepackage[labelfont=bf]{caption}

\usepackage{color}

\newcommand{\style}[1]{\textcolor{blue}{#1}}

\newcolumntype{L}[1]{>{\raggedright\arraybackslash}p{#1}}
\newcolumntype{R}[1]{>{\raggedleft\arraybackslash}p{#1}}
\newcolumntype{C}[1]{>{\centering\let\newline\\\arraybackslash\hspace{0pt}}m{#1}}
\newcolumntype{?}{!{\vrule width 1pt}}
\makeatletter
\newcommand{\thickhline}{%
    \noalign {\ifnum 0=`}\fi \hrule height 1pt
    \futurelet \reserved@a \@xhline
}
\newcolumntype{"}{@{\hskip\tabcolsep\vrule width 1pt\hskip\tabcolsep}}
\makeatother

\title{How to Train Text Summarization Model with Weak Supervisions}

\author{Yanbo Wang$^{1}$ Wenyu Chen$^{1}$ and Shimin Shan$^{3}$}
\date{%
    $^1$ School of Computer Science and Technology, North University of China, Taiyuan\\
    $^2$ School of Semiconductor and Physics, North University of China, Taiyuan\\
}

\begin{document}

\maketitle

\begin{abstract}
Currently, machine learning techniques have seen significant success across various applications. Most of these techniques rely on supervision from human-generated labels or a mixture of noisy and imprecise labels from multiple sources. However, for certain complex tasks, even noisy or inexact labels are unavailable due to the intricacy of the objectives. To tackle this issue, we propose a method that breaks down the complex objective into simpler tasks and generates supervision signals for each one. We then integrate these supervision signals into a manageable form, resulting in a straightforward learning procedure. As a case study, we demonstrate a system used for topic-based summarization. This system leverages rich supervision signals to promote both summarization and topic relevance. Remarkably, we can train the model end-to-end without any labels. Experimental results indicate that our approach performs exceptionally well on the CNN and DailyMail datasets. 
\end{abstract}

\section{Introduction}

Machine learning methods have achieved great success and are widely used in practice. 
Most of these methods are based on supervised learning and rely heavily on a large amount of manually-labeled supervision. 
For reference, the state-of-the-art machine translation model, GNMT~\cite{wu2016google}, is trained on a dataset containing 6M sentence pairs and 340M words, state-of-the-art image classification model, VGGNet~\cite{simonyan2014very}, is trained on a dataset of 1.2M labeled images. 
However, labeling training data has increasingly become the bottleneck for machine learning systems because it is usually expensive, time-consuming and error-prone.

To alleviate this issue, a number of weak supervision methods have been explored. 
For example, a number of weak supervision methods are developed to handle the problem where labels are noisy, incomplete, inaccuracy~\cite{zhou2017brief,karger2011iterative,ratner2018training,du2023abds,lu2018multi,sachan2018learning}. 
Existing works focus on generate labels from multiple sources, such as knowledge base (also known as distant supervision)~\cite{mintz2009distant,de2016deepdive,takamatsu2012reducing}, feature annotation~\cite{mann2010generalized,zaidan2008modeling}, heuristic pattern~\cite{gupta2014improved,hearst1992automatic} and crowd-sourcing noisy labels~\cite{karger2011iterative,zhang2014spectral}. 
~\cite{ratner2017snorkel,ratner2018training,ratner2019role} focus on combining (denoising and combining) labels from different sources. 
However, all these existing methods are based on the assumption that the noisy or inexact labels are available.
The assumption is too restrictive for some tasks, e.g., topic-based summarization. 
Unlike general text summarization that covers all the salient points of a document~\cite{kedzie2018content,see2017get}, topic-based text summarization aims to create short summaries of documents in the context of an topic, Table~\ref{table:demo} provides a simple example to demonstrate what is topic-based summarization. 
The task is quite complex. The objective (goal) is that the generated text has to be not only relevant to the topic but also informative.

In this paper, we decompose the whole objective (goal) into basic and simple tasks and generate supervision signals for each. 
Then, we propose a unified framework to integrate various supervision signals to represent the combined effect. 
Since knowledge has already been encoded into a supervision signal, we don't have to specifically design neural architecture and learning objectives, and the resulting learning and inference procedure is quite simple.

As a case study, we apply our approach on a novel task, topic-based summarization, where it is hard to acquire the labeled training data. 
Thus, it is a more challenging task than general text summarization. 
We decompose the objective of topic-based summarization into two basic requirements: informativeness and relevance.
To encourage informativeness, we use general summary labels as the supervision signals.
On the other hand, to specify the relevance between the topic and the sentence in the source document, we first design a simple rule that checks if the keyword in the topic would appear in the current sentence. 
Second, we use semantic similarity between topics and sentences in documents to further enhance it. 
In addition, supervision signals could also come from a pre-trained model for a correlated task, such as a context Question Answering (QA) model.
The target of Context question answering is to find an answer to the question in a given context, where the answer to each question is a segment of the context. 
If we let the topic be the question and the source document be the context paragraph, we acquire the answer sentence via a pre-trained QA model, which could supervise our task.

We train the model on CNN/DailyMail dataset~\cite{hermann2015teaching} that is usually used for general summarization and evaluate our method on topic-based CNN and DailyMail dataset~\cite{hasselqvist2017query}.  Empirical results demonstrate that on topic-based extractive summarization our method can achieve desirable accuracy without using topic-based reference summary.

Our paper is structured as follows.
Firstly, we briefly review the closely related literature. 
Then, we describe our method, including the framework and how to generate various supervision signals that encode knowledge. 
Then, we demonstrate the empirical procedure and report the experimental results. 
Finally, we conclude our paper.


\begin{table}[]
\centering
\small
\resizebox{\textwidth}{!}{
\begin{tabular}{L{14.1cm}@{}} 
\cmidrule[\heavyrulewidth]{1-1} 
{\bf Source document}\ \ \ \ (cnn) – the United States have named former Germany captain Jurgen Klinsmann as their new national coach, just a day after sacking Bob Bradley. 
Bradley, who took over as coach in January 2007, was relieved of his duties on Thursday, and U.S. soccer federation president Sunil Gulati confirmed in a statement on Friday that his replacement had already been appointed. [...]
\\[4pt]
{\bf topic}\ \ \ \  United States
\\[4pt]
{\bf Ground-truth reference summary }\ \ \ \ \ Jurgen Klinsmann is named as coach of the United States national side.
\\[4pt]
\cmidrule[\heavyrulewidth]{1-1}
\end{tabular}
}
\caption{An example of topic-based summarization. Ground-truth reference summary is usually abstractive. }
\label{table:demo}
\end{table}

\section{Related Studies}
\label{sec:related}

We review some closely related works in this section and discuss their difference with our method. 

\paragraph{topic-based Summarization}
Text summarization is a fundamental task in natural language processing community. 
It can be divided into two paradigms: extractive summarization and abstractive summarization. 
Extractive summarization selects salient sentences from the original text to create a summary~\cite{kedzie2018content,narayan2018ranking,nallapati2017summarunner,arumae2019guiding,liu2019fine}.
In contrast, abstractive summarization learns an internal language representation to generate more human-like summaries, paraphrasing the intent of the original text~\cite{see2017get,chen2018fast,gehrmann2018bottom}. 
In recent years, most of topic based summarization methods~\cite{hasselqvist2017query,nema2017diversity,baumel2018query,krishna2018generating,narayan2018don} are abstractive summarization, with encode-attend-decode framework, which support end-to-end training. 
However, it is totally data-driven and requires a large amount of labeled data. 
In contrast, topic-based extractive summarization are less explored and usually based on conventional machine learning methods instead of deep learning and include manual feature design.
\cite{wang2016query,feigenblat2017unsupervised} cast sentence subset selection problem as a combinatorial optimization problem, where objective encourage both topic-relevance and summarization.  
\cite{li2011generating} infer the topic of sentences via LDA and then select the sentence via ranking and compression. These extractive methods are not based on neural networks and don't achieve SOTA performance.

\paragraph{Weak Supervision}
As machine learning models continue to increase in complexity,  collecting massive hand-labeled training sets is prohibitively expensive and error-prone. 
A bunch of weak supervision methods were designed to fix the issue, where labels come from multiple sources, such as knowledge base (also known as distant supervision)~\cite{mintz2009distant,de2016deepdive,takamatsu2012reducing}, feature annotation~\cite{mann2010generalized,zaidan2008modeling}, pattern-based heuristic~\cite{gupta2014improved,hearst1992automatic} and crowd-sourcing~\cite{karger2011iterative,zhang2014spectral}. 
~\cite{ratner2017snorkel,ratner2018training} focus on combining noisy labels from different sources. 
Concretely, \cite{ratner2017snorkel} denoise and combine several human-generated heuristic label via minimizing average loss over various noisy labels. 
\cite{ratner2018training} developed a novel matrix completion-style problem to recover the truth label from multiple weak supervision sources. 
However, all these existing methods are based on inaccurate, inexact labels. 
Different from them, in this paper, we focus on the task where even noisy labels are unavailable. Specifically, we decompose objective into some simple targets, and generate supervision for each of them.

In addition, \cite{hu2016harnessing,hu2018deep} are also important motivations for our work. 
Unlike most of deep learning models that incorporate knowledge in the design of model architecture, they encode knowledge (such as logic rule or constraint) into loss objective and let neural network encode the knowledge automatically.

Also, our method integrates various supervision signals directly so that we don't have to change the learning objective and design complex model architecture, which makes the learning and inference procedure much simpler than ~\cite{ratner2017snorkel,ratner2018training,hu2016harnessing,hu2018deep}.

\section{Topic-based Summarization with Rich Supervisions}
\label{sec:method}

\subsection{Topic-based Extractive Summarization}
\label{sec:extractive}

In this paper, we focus on topic-based extractive summarization.
The target is to generate an extractive summary of the document with respect to the topic. topic can be several words or a sentence. 
It is quite common for a topic to be an entity name that occurs in the source document. 
Each data sample contains a topic and a document containing $n$ sentences $\sbb_1, \cdots, \sbb_n$. 
It is formulated as a sequence tagging problem with $n$ binary extractive labels $y_1, \cdots, y_n$.

Available reference summaries (denoted $\rbb$) are usually human-generated abstracts. 
A common method is to generate binary extractive labels $y_1, \cdots, y_n$ via automatically aligning human abstracts and source documents~\cite{kedzie2018content}.

\subsection{Learning with Rich Supervisions}
\label{sec:learn}


In label-free scenarios, supervision is regarded as a replacement of labels to guide the learning process. 
In a binary classification problem, each supervision is regarded as ``soft'' relaxation of binary labels, so it ranges from 0 to 1. 
Suppose we have already collected a number of supervisions, denoted $\calY$. 
In the learning procedure, the target is the integration of all the supervisions, representing the combined effect of all supervisions.  
Specifically, the learning target is to minimize the following objective function  
\begin{equation}
\label{eq:obj1}
\begin{aligned}
\calL(\Theta) & = \sum_{i = 1}^{n} \text{Cross-Entropy} \big(p_i, \tilde{y}_i \big) = \sum_{i = 1}^{n} -  \tilde{y}_i \log(p_i) - (1 -  \tilde{y}_i)\log(1 - p_i) ,\\
\end{aligned}
\end{equation}
where $i$ represents $i$-th sentence in source document, $\tilde{y}_i  = \sum_{y \in \calY} \lambda  y_i$ is the integrated supervision. 
hyperparameter $\lambda$s are between 0 and 1, weighing the importance of certain supervision in the whole objective. 
We assume the sum of all $\lambda$s equal to 1 to guarantee $0 \leq \tilde{y}_i\leq 1 $. 
$p_i$, short for $p_{\Theta}(y_i \vert \sbb_1, \cdots \sbb_n, )$, is the predicted probability of the $i$-th sentence.
$\Theta$ are the parameters of the model.

We discuss how to create the \texttt{supervisions} using different ways. 
They are motivated by different properties of topic-based extractive summarization. 
topic-based summarization is a comprehensive task that balances summarization quality and topic-relevance. 
The generated summary need to be not only concise and informative, but also relevant to the topic. 
All the supervisions are motivated by the general idea and can be divided into several categories: (i) labels for other tasks, (ii) rule-based supervision, (iii) semantic similarity (iv) pretrained model (of a related task).  
All of the supervisions are described as follows. 
A brief description is available in Table~\ref{table:alllabel}.

\paragraph{General Summary Labels}\ \\ 
If training data has labels for other tasks, these labels may be helpful. 
topic-based summarization requires the generated summary to be informative and concise. It is a natural idea to incorporate the general reference summary to encourage the ``informativeness'' and ``conciseness''. 
On the other hand, it's much easier to get a general reference summary than a summary based on certain topics.
The reference summaries are usually human abstracts. 
Binary extractive labels can be obtained via aligning human abstracts and source document~\cite{kedzie2018content}. 
Thus, labels for general extractive summarization is used for our task, denoted $y_{\text{1}}^{\text{e}}, \cdots, y_n^{\text{e}} \in \{0,1\}$, corresponding to sentences $\sbb_1, \cdots, \sbb_n$, respectively.

The training corpus we use doesn't have an topic, and we want to generate it. 
In this paper, we use topic CNN and topic DailyMail dataset (as described in Section~\ref{sec:setup}) as a test set, where topic is usually an entity that comes from the source document. 
To make our training data consistent with test data, we extract entities from the source document using the Named Entity Recognition (NER) toolkit based on NLTK\footnote{\url{https://www.nltk.org/}}. 
Then, we focus on the supervision that encourages the relevance between topics and sentences in the document.

\paragraph{Rule-based Supervision}\ \\ 
Supervision can also be generated via simple rules. 
For topic-based summarization, if the keyword topic occurs in some sentences in the source document, then we claim these sentences are more relevant to the topic. 
We define an indicator $y^{\text{a}}_i$ to measure if keyword in topic would appear in the $i$-th sentence, if keyword in topic appear in $\sbb_i$, then $y^{\text{a}}_i  = 1$, otherwise $y^{\text{a}}_i = 0 $.

\paragraph{Semantic Similarity}\ \\ 
We use four semantic metrics to capture similarity: (A) Word Similarity, (B) topic-sentence Similarity, (C) Reference-Sentence Similarity, and (D) Sentence-Sentence Similarity. 
First, we measure the word-level relevance according to similarity between the entity in topic and entity in each sentence, denoted $y^{\text{w}}_i$. 
Suppose $\mathcal{W} =\{w_1, w_2, \cdots \}$ are all entities in sentence $\sbb_{i}$, $\mathcal{V} = \{v_1, v_2, \cdots \}$ are entities in topic, then \textbf{(A) word similarity} $y_i^{\text{w}} $ is defined as 
\begin{equation}
\begin{aligned}
y^{\text{w}}_i = \max_{w\in \mathcal{W}, v\in\mathcal{V}} \max \big( 0, \text{sim}(w, v) \big), \ \ \text{for} \ i = 1, \cdots, n.  
\end{aligned}
\end{equation}
where $\text{sim}(w,v)$ is similarity between word $w$ and $v$, here we use cosine distance of word embedding~\cite{mikolov2013distributed} to measure it.

Now, we want to measure sentence-level relevance. 
Concretely, we use BERT-based sentence embedding to represent topic, sentences in source document ($\sbb_1, \cdots, \sbb_n$), and abstract reference summary ($\rbb$) as a fixed-size vector using BERT-embedding, denoted $\bb_a$, $\{\bb_1, \cdots, \bb_n\}$ and $\bb_r$, respectively. 
The cosine distance between two sentence vectors is used to measure the semantic similarity between two sentences.

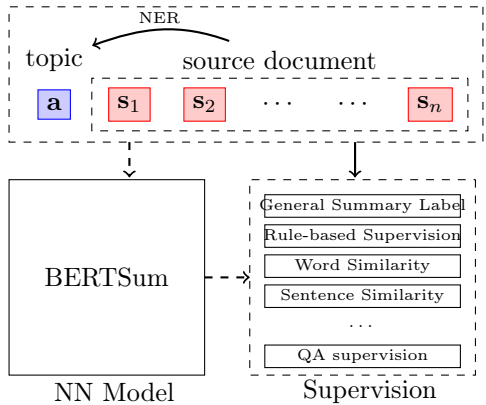
\begin{figure}
\centering
\begin{tikzpicture}[
  hid/.style 2 args={
    rectangle split,
    rectangle split horizontal,
    draw=#2,
    rectangle split parts=#1,
    fill=#2!20,
    outer sep=1mm}]
\node[hid={1}{red}, align=center] (s1) at (0,3.5) { $\mathbf{s}_1$};
\node[hid={1}{red}, align=center] (s2) at (1.0,3.5) {  $\mathbf{s}_2$};
\node at (2,3.5) (cdot) {$\cdots $};
\node at (3,3.5) (cdot2) {$\cdots $};
\node[hid={1}{red}, align=center] (s2) at (4.0,3.5) {  $\mathbf{s}_n$};
\draw[style=dashed] (4.5, 3.85) rectangle (-0.5, 3.15); 
\draw[style=dashed] (4.7, 4.8) rectangle (-1.6, 3); 
\node at (2,4.1) (sourcedoc) {source document};
\node[hid={1}{blue}, align=center] (r0) at (-1,3.5) { $\mathbf{a}$};
\node at (-1,4.1) (topic) {topic};
\draw [->,thick] (sourcedoc) to [in = 20, out = 160] (topic);
\node[font=\fontsize{6}{7}\selectfont] at (0.4,4.65) (ner) {NER};
\draw[] (1, 2.5) rectangle (-1.6, -0.1) ; 
\node at (-0.3,1.2) { BERTSum }; 
\draw[style=dashed] (4.6, 2.5) rectangle (1.6, -0.1) ; 
\node at (3.2, -0.32) {Supervision};
\node at (-0.2, -0.32) {NN Model};
\draw[->,thick,style=dashed] (1,1.2) to (1.6,1.2);
\draw[->,thick,style=dashed] (0,3) to (0,2.53);
\draw[->,thick] (3,3) to (3,2.53);
\draw[] (4.4, 2.3) rectangle (1.8, 2.0) ; 
\node[font=\fontsize{6}{6}\selectfont] at (3.1, 2.15) {General Summary Label};
\draw[] (4.4, 1.9) rectangle (1.8, 1.6) ; 
\node[font=\fontsize{6}{6}\selectfont] at (3.1, 1.75) {Rule-based Supervision};
\draw[] (4.4, 1.5) rectangle (1.8, 1.2) ; 
\node[font=\fontsize{6}{6}\selectfont] at (3.1, 1.35) {Word Similarity}; 
\draw[] (4.4, 1.1) rectangle (1.8, 0.8) ; 
\node[font=\fontsize{6}{6}\selectfont] at (3.1, 0.95) { Sentence Similarity};
\node[font=\fontsize{6}{6}\selectfont] at (3.1, 0.55) { $\cdots $};
\draw[] (4.4, 0.3) rectangle (1.8, 0.0) ; 
\node[font=\fontsize{6}{6}\selectfont] at (3.1, 0.15) {QA supervision};
\end{tikzpicture}
\caption{
The framework of this paper. 
The dashed line represents the learning procedure, and the solid line represents supervision generation. 
topic is extracted from the source document using NER. 
We don't modify neural architecture and regard it as a black box. 
Various supervisions are integrated as the learning target.  
} 
\label{fig:M1}
\end{figure}

Second, \textbf{(B) topic-Sentence Similarity} $y_i^{\text{as}}$ is defined to measured the relevance between topic  and $i$-th sentence $\sbb_{i}$, 
\begin{equation}
\label{eq:y4}
\begin{aligned}
y^{\text{as}}_i & = \max \big( 0, \text{cos}(\bb_{}, \bb_i) \big), \ \ \text{for} \ i = 1, \cdots, n, \\ 
\end{aligned}
\end{equation}
where $\text{cos}(\cdot, \cdot)$ represent the cosine similarity between two vectors. 

Third, \textbf{ (C) Rerefence-Sentence Similarity}
 $y_i^{\text{rs}}$ is defined to measured the relevance between human-generated abstract $\rbb$ and $i$-th sentence $\sbb_{i}$, 
\begin{equation}
\label{eq:y5}
\begin{aligned}
y^{\text{rs}}_i & = \max \big( 0, \text{cos}(\bb_{}, \bb_i) \big), \ \ \text{for} \ i = 1, \cdots, n. \\ 
\end{aligned}
\end{equation}
It serves as a complementary for binary extractive labels $y^{\text{e}}_i \in \{0, 1\}$.

Last, the general intuition is that the sentence that has higher similarity with other sentences in the document is more informative and more likely to be selected in summary. 
We use $t_{i,j}$ to denote the similarity between $i$-th and $j$-th sentence.,
\begin{equation}
\label{eq:y61}
\begin{aligned}
t_{i,j} = \max \big( 0, \text{cos}( \bb_i, \bb_j)\big),
\end{aligned}
\end{equation}
\textbf{(D) Sentence-Sentence Similarity} $y^{\text{ss}}_i$ is defined as 
\begin{equation}
\label{eq:y62}
\begin{aligned}
y^{\text{ss}}_i = \frac{1}{n-1}\sum_{j \in \mathcal{S}^{-i} } t_{i,j}, 
\end{aligned}
\end{equation}
where $\mathcal{S}^{-i}$ denotes the set that remove $i$ from $\{1,\cdots,n\}$.

\paragraph{Pretrained model: Question Answering (QA) Supervision}
\label{sec:qa}\ \\
Supervision can also come from pre-trained model from related task. 
Question Answering on SQuAD  dataset~\cite{rajpurkar2016squad} is a task to find an answer on question in a given context (e.g, paragraph from Wikipedia), where the answer to each question is a segment of the context. 
This task is similar to topic-based summarization, where topic  can be seen as the question, documents correspond to context paragraph. Generated summary correspond to answer. 
Thus, we directly input our data (topic  and sentences in source document $\sbb_1, \cdots, \sbb_n$) into the well-trained question answering model trained on SQuAD dataset.
We use the pre-trained model available at \url{http://docs.deeppavlov.ai/en/latest/components/squad.html}. 
The output answer is regarded as generated summary. 
Here the generated summary is regarded as human-generated abstract. 
By aligning generated summary and source document~\cite{kedzie2018content}, we generate supervisions for each sentence in source document, denoted $  y^{\text{qa}}_1, \cdots, y^{\text{qa}}_n \in  \{0, 1\} $.

\begin{table}[t]
\small 
\centering
\resizebox{\textwidth}{!}{
\begin{tabular}{l l l} 
\cmidrule[\heavyrulewidth]{1-3}
Class & Supervisions & Short Explanation \\ \cmidrule{1-3}
Labels from other tasks &  General Summary Label  & Binary labels for extractive summarization. \\
Rule-based & topic Indicator & If key word in topic occur in sentence. \\ 
\multirow{4}{*}{Semantic Similarity} & Word Similarity &  Similarity of keyword between topic and sentences. \\
 & topic-Sentence Similarity & Between topic and sentences. \\
 & Reference-Sentence Similarity & Between general summary reference and sentence. \\ 
 & Sentence-Sentence Similarity & Between sentences in document.\\ 
Pre-trained Model & QA induced supervision & Supervision generated from QA model.  \\ 
\cmidrule[\heavyrulewidth]{1-3}
\end{tabular}}
\caption{All supervisions. }
\label{table:alllabel}
\end{table}

\section{Experiment}
\label{sec:experiment}

In this section, we describe the empirical evaluation of our method. First, we introduce the datasets we use. 

\subsection{Experiment Setup}
\label{sec:setup}

We use three datasets as follow. 
First, CNN-DailyMail (CNN-DM) is a standard corpus for general text summarization~\cite{hermann2015teaching}. 
It contains online news articles (781 tokens on average) paired with multi-sentence summaries (3.75 sentences or 56 tokens on average). 
The other two corpus are topic-based, topic CNN (A-CNN) and topic DailyMail (A-DM)~\cite{hasselqvist2017query}. 
Each article corresponds to a number of human-written highlights, which summarize different topics of the article. 
Each summarization contains one sentence (14.5 tokens on average). 
These corpus are a mix of news on different topics including politics, sports, and entertainment. 
The statistics of these datasets are described in Table~\ref{table:dataset}.

In our method, during training procedure,  we use training set of CNN-DM, and use test set of A-CNN and A-DM for testing. 
It is also worth mentioning that we guarantee that the test set of A-CNN and A-DM do not occur in CNN-DM training set.

Sentences are split by CoreNLP. We follow the preprocessing  method described in \cite{see2017get} for CNN-DM. For A-CNN and A-DM, we follow the preprocessing method described in \cite{hasselqvist2017query}\footnote{\url{https://github.com/helmertz/querysum-data/}}.

\begin{table}[t]
\small
\centering
\begin{tabular}{r l l l} 
\cmidrule[\heavyrulewidth]{1-4}
& {\bf Train doc/topic} & {\bf Valid doc/topic} & {\bf Test doc/topic} \\ \cmidrule{1-4}
CNN-DM & 287K/- & 13K/- & 11K/- \\
topic CNN & 89K/284K & 1.4K/4.4K & 0.7K/2.2K \\ 
topic DM & 212K/784K & 3.3K/11.9K & 3.3K/12.2K   \\ 
\cmidrule[\heavyrulewidth]{1-4}
\end{tabular}
\caption{Data Statistics.}
\label{table:dataset}
\end{table}

The baseline methods include
\begin{itemize}
\item \textbf{Oracle}. To see the accuracy ceiling of topic-based extractive summarization, we select the sentences according to extractive labels. That is to say, the oracle method reaches approximately the maximum possible accuracy for extractive method on this task. 
\item \textbf{BERTSum}. BERTSum~\cite{liu2019fine} achieved state-of-the-art performance on extractive summarization. Here, we evaluate pre-trained BERTSum model (trained on CNN-DM for extractive summarization) on our task. 
\end{itemize}

BERTSum\footnote{Code is publicly available at \url{https://github.com/nlpyang/BertSum}. } achieved state-of-the-art performance on extractive summarization thanks to pre-trained BERT initialization~\cite{liu2019fine}. 
our model is based on BERTSum and use the same neural architecture with the same learning rate schedule. The topic is added at the beginning of document and is regarded as a single sentence. 
All models are trained for 200,000 iterations on a Titan X GPU. 
During testing, we rank all the checkpoints according to their losses on the validations set, choose the top-3 ones, and report the averaged results on the test set. 
Regarding hyperparameter, we set all the hyperparameter $\lambda$ equal to each other.



For extractive summarization, there are usually constraints on generated summary. For example, in \cite{kedzie2018content}, the generated summary has at most 100 words. 
In \cite{liu2019fine}, the generated summary has at most 3 sentences.  Here the reference summary has one sentence and average 15 tokens. We constrain the generated summary to one sentence or 20 words, and report the performance for both cases.

When predicting summaries for a new document and corresponding topic, we first use the models to obtain the score for each sentence. 
We then rank these sentences by the scores from higher to lower. 
Summaries are generated using by selecting top-1 sentence or first-20 words. We report performance for both cases.

The generated summaries are compared with the ground truth summary. 
ROUGE scores~\cite{lin2004rouge} are standard metrics to measure the quality of summaries. 
We report $F_1$ score of ROUGE-1, ROUGE-2 and ROUGE-L in results.

\subsection{Results}
\label{sec:results}

In this section, we report and analyze the experimental results.

\paragraph{Comparison with Baseline}\ \\ 
The results for baseline methods and the variants of our methods on both A-CNN and A-DM are reported in Table~\ref{table:basecnn} and \ref{table:basedm}, respectively. 
Compared with BERTSum that is trained on general summarization dataset, topic-BERTSum can significantly improve the accuracy, validating the effectiveness of the neural architecture (adding topic at the beginning of document and regard it as a single sentence).

\paragraph{Add Supervisions Incrementally}\ \\ 
First, we incrementally add different kind of supervisions, and observe whether it can improve the accuracy. 
Specifically, we show the results for a series of experiment:
``\texttt{ext-label}'' (exactly \texttt{BERTSum}); 
``\texttt{ext-label \& rule-based}'';
``\texttt{ext-label \& rule-based \& sem-sim}'' (i.e., ``\texttt{all -- $\{\text{QA}\}$}'', contains all these supervisions except QA induced supervision, where ``sem-sim'' is the short for semantic similarity) and ``\texttt{all}'' (contains all of the supervisions). We find that the accuracy increase significantly as we incorporate more supervisions.  

\paragraph{Effect of each Supervision}\ \\ 
Second, since the optimal setting is all these supervisions, we remove each of these 4 supervisions, and observe the change in accuracy. 
The combination include 
``\texttt{all}'' (contains all of the supervisions), 
``\texttt{all -- $\{\text{sem-sim}\}$}'' (contains all these supervisions except semantic similarity supervision); 
``\texttt{all -- $\{\text{rule-based}\}$}'' (contains all these supervisions except rule-based supervision), 
``\texttt{all -- $\{\text{ext-label}\}$}'' (contains all these supervisions except general summary label); 
``\texttt{all -- $\{\text{QA}\}$}'' (contains all these supervisions except QA induced supervision).   
By observing results, we find that all of the supervisions are helpful on topic-based summarization. 
Among all of the supervisions, rule-based supervision and general summary label are most important supervisions for the task.

\paragraph{Case study}\ \\ 
Also, we show an example in Table~\ref{table:example}. 
We can find that if we don't include topic relevance supervision, the generated summary would be close to general summary. In contrast, if we don't include general summary supervision, the generated summary would only be topic-related, always involves some details. 
The model trained on all supervisions will produce the most correct answer.

\begin{table}[t]
\centering
\small
\resizebox{\textwidth}{!}{
\begin{tabular}{@{}l r r r r r r@{}}
\cmidrule[\heavyrulewidth]{1-7}
\multirow{2}{*}{Model} & \multicolumn{3}{c}{{\bf1 sentence}} & \multicolumn{3}{c}{{\bf20 words}} \\ 
 & {\bf Rouge-1} & {\bf Rouge-2} & {\bf ROUGE-L} & {\bf Rouge-1} & {\bf Rouge-2} & {\bf ROUGE-L} \\
\cmidrule[\heavyrulewidth]{1-7}
topic-BERTSum  & 27.87 & 13.08 & 23.83 & 27.01 & 12.35 & 24.11 \\ 
BERTSum (ext-label) & 18.32 & 6.28 & 15.37 & 17.98 & 6.24 & 15.97 \\ 
\cmidrule{1-7}\morecmidrules\cmidrule{1-7} 
ext-label \& rule-based & 24.48 & 9.73 & 21.01 & 24.63 & 9.98 & 21.94 \\ 
all--$\{\text{QA}\}$\footnote{ext-label \& rule-based \& sem-sim }  & 26.35 & 11.73 & 22.23 & 26.12 & 11.87 & 23.09 \\  
all--$\{\text{sem-sim}\}$ & 26.64 & 11.97 & 22.84 & 26.38 & 11.83 & 23.40   \\ 
all--$\{\text{ext-label}\}$  & 25.53 & 10.79 & 21.48 & 25.16 & 10.78 & 22.02   \\ 
all & \textbf{27.73} & \textbf{12.78} & \textbf{23.62} &  \textbf{27.27} & \textbf{12.66} & \textbf{24.16}  \\ 
\cmidrule{1-7}\morecmidrules\cmidrule{1-7}
ORACLE  & 34.55 & 18.81 & 30.34 & 33.28 & 18.43 & 30.81   \\ 
\cmidrule[\heavyrulewidth]{1-7}
\end{tabular}}
\caption{Results of topic based extractive summarization for all methods on \textbf{topic-CNN} dataset. 
We report ROUGE (\%) $F_1$-score of our model (with different settings) and baseline model.}
\label{table:basecnn}
\end{table}

\begin{table}[t]
\centering
\small
\resizebox{\textwidth}{!}{
\begin{tabular}{@{}l r r r r r r@{}}
\cmidrule[\heavyrulewidth]{1-7}
\multirow{2}{*}{Model} & \multicolumn{3}{c}{{\bf1 sentence}} & \multicolumn{3}{c}{{\bf20 words}} \\ 
 & {\bf Rouge-1} & {\bf Rouge-2} & {\bf ROUGE-L} & {\bf Rouge-1} & {\bf Rouge-2} & {\bf ROUGE-L} \\
\cmidrule{1-7}
BERTSum (ext-label) & 19.92 & 8.04 & 17.43 & 19.03 & 6.98 & 16.39 \\ 
\cmidrule{1-7}\morecmidrules\cmidrule{1-7} 
ext-label \& rule-based & 27.13 & 13.83 & 23.81 & 26.83 & 13.01 & 23.98 \\ 
all--$\{\text{QA}\}$\footnote{ext-label \& rule-based \& sem-sim }  & 29.61 & 14.99 & 15.03 & 29.45 & 14.23 & 25.45  \\  
all--$\{\text{sem-sim}\}$  & 29.91 & 15.90 & 25.93 &  29.78 & 14.98 & 26.25    \\ 
all--$\{\text{ext-label}\}$ & 28.87 & 14.51 & 24.59 & 29.20 & 14.03 & 25.43  \\ 
all & \textbf{30.75} & \textbf{16.13} & \textbf{26.40} & \textbf{30.53} & \textbf{15.42} & \textbf{26.81}  \\ 
\cmidrule{1-7}\morecmidrules\cmidrule{1-7}
ORACLE  & 37.32 & 23.52 & 33.41 & 35.50 & 21.41 & 31.84  \\
\cmidrule[\heavyrulewidth]{1-7}
\end{tabular}}
\caption{Results of topic-based extractive summarization for all methods on \textbf{topic-DailyMail} dataset. 
We report ROUGE (\%) $F_1$-score of our model (with different settings) and baseline model. }
\label{table:basedm}
\end{table}

\begin{figure}[t]
\centering
\resizebox{0.9\columnwidth}{!}{
\begin{tabular}{ccc}
\hspace{-3mm}\includegraphics[width=6cm]{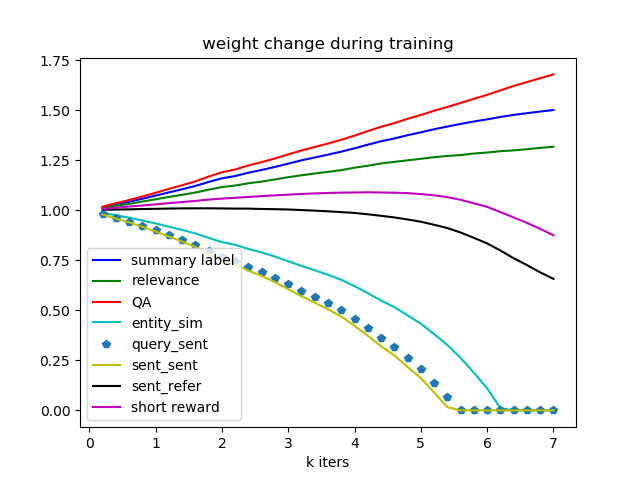} & \hspace{-2mm}\includegraphics[width=6cm]{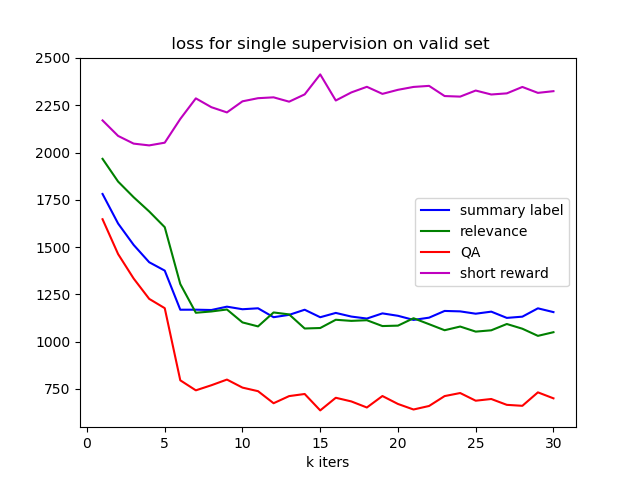} & \hspace{-2mm}\includegraphics[width=6cm]{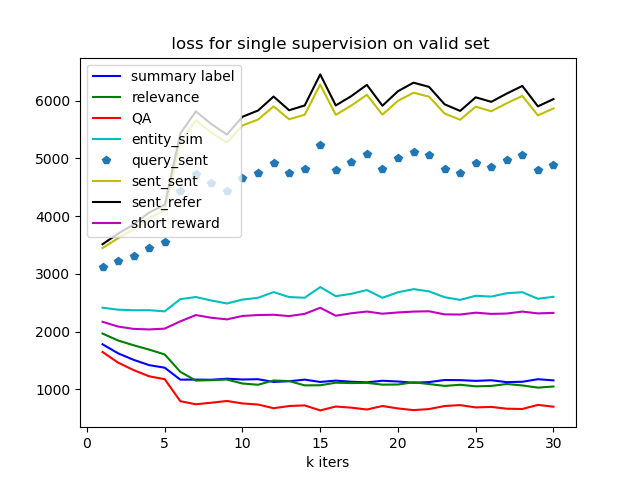} \\
\hspace{-3mm}(a)weight change for ``all''-supervisions & (b) 4 supervision losses for ``all''-supervisions & (c) same as (b), but 8 losses \\
\hspace{-3mm}\includegraphics[width=6cm]{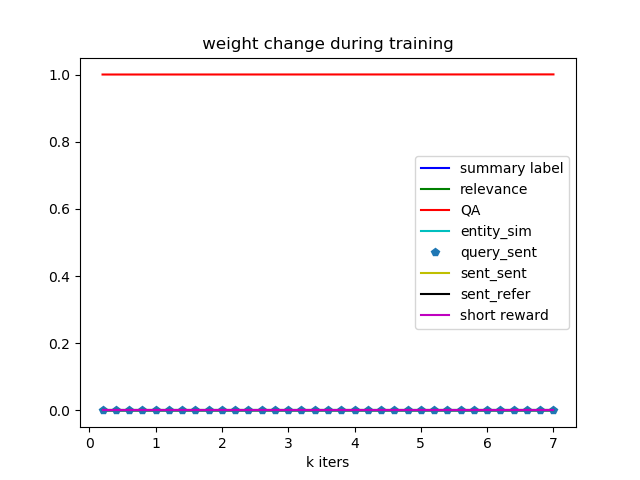} & \hspace{-2mm}\includegraphics[width=6cm]{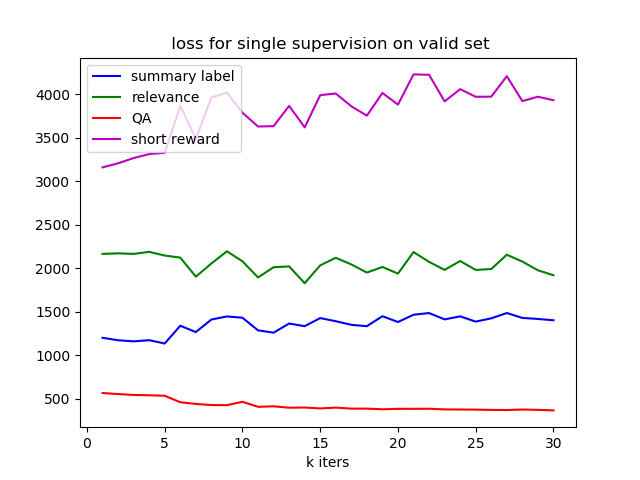} & \hspace{-2mm}\includegraphics[width=6cm]{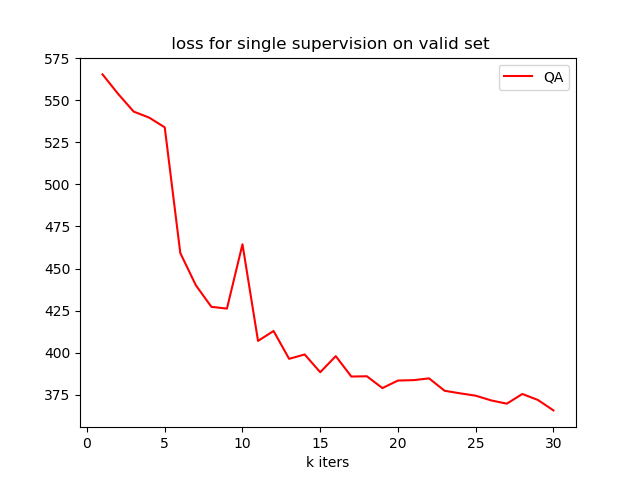} \\
\hspace{-3mm}(d)weight change for ``only QA'' & (e) 4 supervision losses for ``only QA'' & (c) same as (b), but only 1 losses \\

\end{tabular}}
\caption{weight and loss change over iterations. Note that weight are fixed after 7k iterations. }
\label{fig:unlabeled}
\end{figure}

\begin{table}[t]
\centering
\small
\resizebox{\textwidth}{!}{
\begin{tabular}{L{14.1cm}@{}} 
\cmidrule[\heavyrulewidth]{1-1} 
{\bf source document}\ \ \ \ Patrick Vieira's move to Manchester City appears to have moved a step closer after Inter Milan coach Jose Mourinho confirmed he has played his last game for the Italian club. 
English Premier League side City had been linked with a move for the 33-year-old midfielder who has a stop-start career at the San Siro since his move from Juventus.
Vieria played in inter's 1-0 win over Chievo and Mourinho paid tribute to his contribution to the club after the match and confirmed his impending departure.
``In particular Vieira was great in his last game for us.
He is a player that we will certainly miss now that he is leaving , '' Mourinho told reporters.
``it was the best way to say goodbye to us and i wish him all the best in his new life.
[...]
Atletico Madrid are closing in on a move for Juventus midfielder Tiago who is set to move to the Spanish La Liga club on loan until the end of the season. \\[4pt]
{\bf topic}\ \ \ City 
\\[4pt]
{\bf extractive summary (all supervisions)}\ \ \ \ 
Patrick Vieira 's move to Manchester city appears to have moved a step closer after Inter Milan coach Jose Mourinho confirmed he has played his last game for the Italian club.
\\[4pt]
{\bf extractive summary (all supervisions -- $\{ \text{ext-label} \}$)}\ \ \ \ English Premier League side City had been linked with a move for the 33-year-old midfielder who has a stop-start career at the San Siro since his move from Juventus.
\\[4pt]
{\bf extractive summary (all supervisions -- $\{\text{semantic similarity}\}$)}\ \ \ \ Vieria played in inter's 1-0 win over Chievo and Mourinho paid tribute to his contribution to the club after the match and confirmed his impending departure. 
\\[4pt]
{\bf ground-truth extractive summary}\ \ \ \ \ Patrick Vieira's move to Manchester City appears to have moved a step closer after Inter Milan coach Jose Mourinho confirmed he has played his last game for the Italian club. 
\\[4pt]
{\bf ground-truth abstractive summary}\ \ \ \ \ 
Patrick Vieira's move to Manchester City appears to have moved a step closer according to Inter Milan coach Jose Mourinho. 
\\[4pt]
\cmidrule{1-1}\morecmidrules\cmidrule{1-1}
{\bf source document}\ \ \ \ Under an almost cloudless sky, family members gathered and soldiers marched in full military dress. Taps echoed in the wind. A wreath of red, white and blue flowers was placed on a grave.
It is a solemn ritual repeated multiple times daily, year-round at Arlington National Cemetery outside Washington. But this ceremony on Tuesday at the resting place of Army Pvt. William Christman carried particular significance.
Christman, a civil war soldier, was the first to be buried at Arlington and the graveside remembrance was held to mark the start of the cemetery's 150th anniversary commemoration, which will continue through June 16. 
$\bf{\cdots \cdots} $
The initial property belonged to George Washington's extended family and then to Robert E. Lee, who left it at the start of the Civil War. Federal troops used it as an encampment, and the federal government purchased 200 acres in 1864 and established a cemetery.
[...]
 \\[4pt]
{\bf topic}\ \ \ George Washington 
\\[4pt]
{\bf extractive summary (all supervisions)}\ \ \ \ 
The initial property belonged to George Washington's extended family and then to Robert E. Lee, who left it at the start of the Civil War. 
\\[4pt]
{\bf extractive summary (all supervisions -- $\{ \text{ext-label} \}$)}\ \ \ \ It is a solemn ritual repeated multiple times daily, year-round at Arlington National Cemetery outside Washington. 
\\[4pt]
{\bf extractive summary (all supervisions -- $\{\text{semantic similarity}\}$)}\ \ \ \ Christman, a civil war soldier, was the first to be buried at Arlington and the graveside remembrance was held to mark the start of the cemetery's 150th anniversary commemoration, which will continue through June 16.
\\[4pt]
{\bf ground-truth extractive summary}\ \ \ \ \ The initial property belonged to George Washington's extended family and then to Robert E. Lee, who left it at the start of the Civil War. 
\\[4pt]
{\bf ground-truth abstractive summary}\ \ \ \ \ 
Property was owned by George Washington's family, Robert E Lee.\\[4pt]
\cmidrule[\heavyrulewidth]{1-1}
\end{tabular}}
\caption{Case study: two examples to compare the ground-truth summary with the generated summary for different models. Extractive summaries are limited to 1 sentence. For example, in first data sample, the source document mainly talks about Patrick Vieira, the topic of interest is Manchester City. If we don't use general summary labels, the generated summaries are usually only topic-related. But if we don't add semantic similarity supervision, the generated summary may not be related to the topic. The model trained on all supervisions will produce the most correct answer. 
Similar things can be found in the second data sample. }
\label{table:example}
\end{table}

\section{Conclusion and Future Work}
\label{sec:conclusion}

In this paper, we restrict on topic-based extractive summarization task.
We have proposed a novel framework that can use a pre-trained NLP model to acquire various supervisions so that our method doesn't need labeled data for this task. Specifically, our model uses general reference summary, word-level relevance (mainly induced by word2vec), sentence-level relevance (induced by BERT-based sentence embedding), and QA-induced information (a well-trained QA model on SQuAD) to get the best performance. 
The empirical results show that the proposed method can achieve desirable accuracy compared with state-of-the-art methods. 
Regarding future work, we plan to explore this general idea in other NLP tasks. 

\section{Future Work}
In this paper, we validate the effectiveness of our idea in NLP. 
Future work could expand the current work in multiple scientific domains, e.g., computer vision~\cite{yi2018enhance,lu2023deep}, gene expression estimation~\cite{chen2021data,lu2019integrated}, multi-omics data integration~\cite{lu2021cot,lu2022cot}, target identification~\cite{zhang2021ddn2,fu2024ddn3}, drug discovery~\cite{wang2024twin,lu2024drugclip}, clinical trial management~\cite{lu2024uncertainty,chen2024uncertainty,chen2024trialbench}, and phenotype prediction~\cite{xu2024mambacapsule}.

\bibliographystyle{abbrvnat}
\bibliography{ref}

\end{document}